# CLEVERarm: A Novel Exoskeleton for Rehabilitation of Upper Limb Impairments

Rana Soltani-Zarrin, Amin Zeiaee, Andrew Eib, Reza Langari, Reza Tafreshi

*Abstract*—CLEVERarm (**C**ompact, **L**ow-weight, **E**rgonomic, **V**irtual/Augmented Reality **E**nhanced **R**ehabilitation **arm**) is a novel exoskeleton with eight degrees of freedom supporting the motion of shoulder girdle, glenohumeral joint, elbow and wrist. Of the exoskeleton's eight degrees of freedom, six are active and the two degrees of freedom supporting the motion of wrist are passive. This paper briefly outlines the design of CLEVERarm and its control architectures.

## I. INTRODUCTION

Stroke affects an increasing portion of the aging population of world, leaving many of the survivors with different levels and forms of disability. There is a recent surge in use of robotic systems for rehabilitation purposes due to their inherent capabilities in producing high intensity, repeatable, and precisely controllable motions [1]. End effector based systems [2] and exoskeletons [3] are the two category of the robotic systems designed to provide automated therapy to stroke patients.

CLEVERarm is a novel upper-limb exoskeleton for rehabilitation of stroke patients. CLEVER ARM has six active, and two passive degrees of freedom (DoF), allowing the motion of shoulder girdle, glenohumeral (GH) joint, elbow, and wrist [4]. An active degree of freedom is used for assisting Flexion/Extension of the elbow, while the remaining five active degrees of freedom are used in the design of the device shoulder to improve the ergonomics of the device. The device is also equipped with visual technologies such as virtual and augmented reality to enable diverse, task specific and immersive training scenarios.

Compactness and weight reduction are two main criteria in development of CLEVERarm. A combination of 3D printed carbon-fiber reinforced plastic and machined aluminum was used to achieve a low weight design. This manuscript briefly reviews the techniques used to ensure viability of using a metal-plastic structure as the chassis of a robotic device considering non-isotropic properties of carbon-fiber reinforced plastic. Additionally, this paper provides brief details on kinematic structure of CLEVERarm, the embodiment of the design and the control architecture.

## II. CLEVER ARM

CLEVER ARM has eight degrees of freedom supporting the motion of shoulder girdle, GH joint, elbow, and wrist. Fig. 1 shows the CAD model of the exoskeleton on a dummy human model, and the actual prototype:

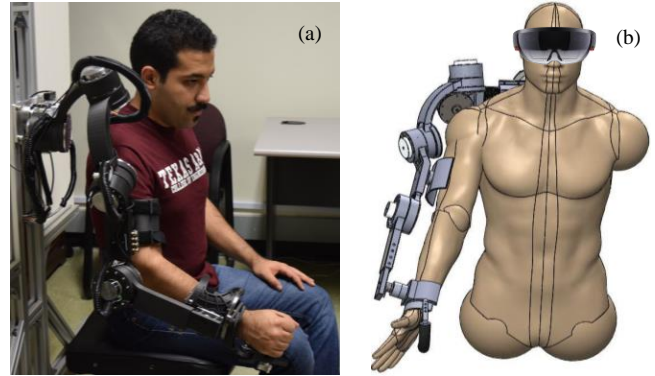

Figure 1 CLEVERarm: (a) prototype, (b) CAD model

The motion of the GH joint and the inner shoulder are supported by five degrees of freedom in the design of CLEVER ARM. Three hinged joints, constituting a spherical linkage, is used to provide the three DoF required for the motion of GH joint. CLEVER ARM uses two active degrees of freedom (a revolute joint followed by a prismatic joint) to model the displacement of GH joint center in the frontal plane of human body which allows accurate tracking of GH joint center path on the frontal plane without approximating it as a circular path [4].

Denavit-Hartenberg (DH) convention was used for kinematic analysis. Fig. 2 shows the assignment of coordinate systems and the corresponding DH parameters where $p_1$ through $p_6$ are the physical parameters of system. While $p_1$ and $p_2$ are constants, other parameters can be changed to accommodate different patient body dimensions.

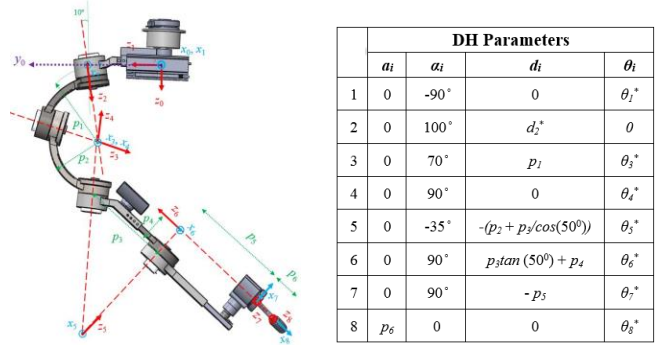

| | **DH Parameters** | | | |
|---|---|---|---|---|
| | $a_i$ | $\alpha_i$ | $d_i$ | $\theta_i$ |
| 1 | 0 | -90° | 0 | $\theta_1^*$ |
| 2 | 0 | 100° | $d_2^*$ | 0 |
| 3 | 0 | 70° | $p_1$ | $\theta_3^*$ |
| 4 | 0 | 90° | 0 | $\theta_4^*$ |
| 5 | 0 | -35° | $-(p_2 + p_3/cos(50^0))$ | $\theta_5^*$ |
| 6 | 0 | 90° | $p_3 tan(50^0) + p_4$ | $\theta_6^*$ |
| 7 | 0 | 90° | $-p_5$ | $\theta_7^*$ |
| 8 | $p_6$ | 0 | 0 | $\theta_8^*$ |

Figure 2. (a) DH coordinate frame, (b) DH parameters

## III. PROTOTYPE DESIGN AND FABRICATION

Weight reduction and compactness of the device are the two criteria in choosing components for motorization of design. Electric motors coupled with strain wave gears (Harmonic Drive LLC), were used in rotary joint while the prismatic joint in the design of exoskeleton was realized by a direct drive linear actuator. To ensure reliability of joint level feedback, an incremental and an absolute encoder is used in each joint. Moreover, the two physical interfaces between the device and body are equipped with 6 axis force/torque sensors to enable achieving back-drivability.

To minimize the weight of the device body, various choices of material and manufacturing techniques were studied and a combination of Aluminum and 3D printed Carbon Fiber (CF) reinforced plastic were selected. While carbon fiber possess many advantageous properties, design of fasteners for metal/composite interfaces and ensuring sufficient rigidity in the structure is challenging due to the anisotropic properties of the composite materials. To address the aforementioned challenges, extensive finite element analysis (FEA) were done to ensure that maximum deflection of the structure in worst case loading scenarios is within the acceptable range. Figure 3.a shows the achieved minimal deflection under worst-case static loading and figure 3.b shows the orientation of the laid carbon fibers achieving the desired strength properties.

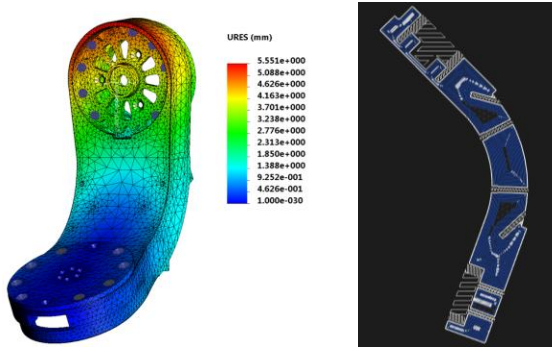

Figure 3. (a) FEA results Max Deflection, (b) Direction of the Carbon Fibers shown in blue in Eiger Software (Markforged Inc.)

Plastic/metal interfaces were designed by distributing the load on larger surfaces to avoid concentration of stress. The weight of the device body is 14 lbs., while the weight of the entire device is 27 lbs.

## IV. CONTROL ARCHITECTURE

The overall control architecture of CELEVRarm is shown in figure 4. This control block diagram is implemented on National Instruments' CompactRio (cRio) RealTime target and FPGA. The main control loop runs at 10 kHz while the force sensors data is acquired by the FPGA at 50 kHz rate. The gravitational model used to cancel the weight of exoskeleton is derived using the inertial properties of the system CAD model (SolidWorks, Dassault Systèmes) [5].

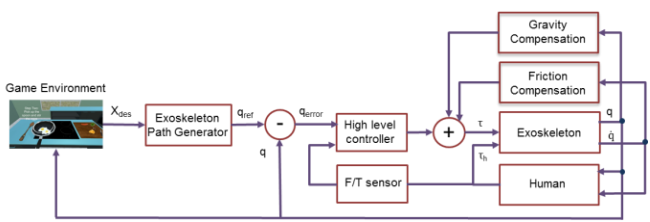

Figure 4. CLEVERarm Control Architecture

Game environments are part of the control architecture of the CLEVERarm since they represent the final desired position for the patient hand. Using the output of the game environment, reference generation block within the control architecture uses the algorithms developed by the authors for generating human-like motions considering the scapulohumeral rhythms [6,7]. Finally, friction compensation is achieved by admittance-based control. The interaction forces measured with the F/T sensors and the desired impedances are used to calculate the desired velocity for the interaction ports [8]. Using the system Jacobian, these desired angular velocities are derived and tracked with a PD control law [9]. Figure 5 shows two different impedance values rendered by the exoskeleton.

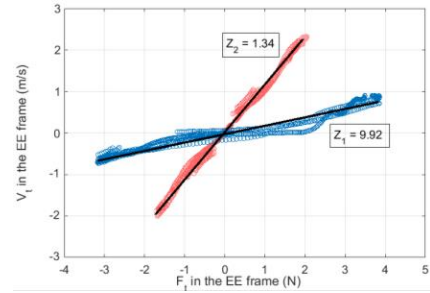

Figure 5. Rendered Impedances by CLEVERarm

## V. CONCLUSION

This paper details the design process and features of the CLEVER ARM, a compact and low weight upper-limb exoskeleton for rehabilitation of stroke patients.